\documentclass[runningheads]{llncs}
\usepackage[T1]{fontenc}
\usepackage{graphicx}
\usepackage{amsmath,amssymb,amsfonts}
\usepackage{multirow}
\usepackage{float}
\usepackage{subcaption}
\usepackage{tcolorbox}
\usepackage{hyperref}
\usepackage{listings}
\usepackage{booktabs}
\usepackage{csquotes}
\usepackage[colorinlistoftodos,prependcaption,textsize=scriptsize]{todonotes}
\begin{document}
\title{Reasoning-Grounded Natural Language Explanations for Language Models}
%
%
\author{Vojtech Cahlik \and
Rodrigo Alves \and
Pavel Kordik}
\authorrunning{V. Cahlik et al.}
%
\institute{Faculty of Information Technology, CTU in Prague, Prague, Czech Republic\\
\email{\{vojtech.cahlik,rodrigo.alves,pavel.kordik\}@fit.cvut.cz}}
\maketitle              
\begin{abstract}
We propose a large language model explainability technique for obtaining faithful natural language explanations by grounding the explanations in a reasoning process. When converted to a sequence of tokens, the outputs of the reasoning process can become part of the model context and later be decoded to natural language as the model produces either the final answer or the explanation. To improve the faithfulness of the explanations, we propose to use a joint predict-explain approach, in which the answers and explanations are inferred directly from the reasoning sequence, without the explanations being dependent on the answers and vice versa. We demonstrate the plausibility of the proposed technique by achieving a high alignment between answers and explanations in several problem domains, observing that language models often simply copy the partial decisions from the reasoning sequence into the final answers or explanations. Furthermore, we show that the proposed use of reasoning can also improve the quality of the answers.

\end{abstract}
%
%
%


\section{Introduction} \label{sec:introduction}

Today's prevalent large language model (LLM) explainability techniques lack the expressivity of natural language, as the explanations are limited in detail and hard to interpret for an untrained user \cite{zhao2024explainability}. On the other hand, natural language explanations \cite{cambria2023survey} can potentially be easy to follow and unlimited in expressivity, but their faithfulness is typically questionable, such as with the simple \textit{answer-then-explain} setting which tends to lead models into fabulating their explanations. Moreover, it is even questionable whether LLMs produce their outputs in a thought process that is anyhow related to human reasoning, as they are in essence mere enhancements of traditional n-gram models \cite{russell2019human}. Chain-of-thought reasoning is one notable improvement of the decision process as the answers tend to follow from the preceding natural language reasoning sequences, but it is too computationally intensive for ubiquitous use.

\begin{figure*}[b!]
  \centering
  \makebox[\textwidth][c]{
    \includegraphics[width=1.5\textwidth]{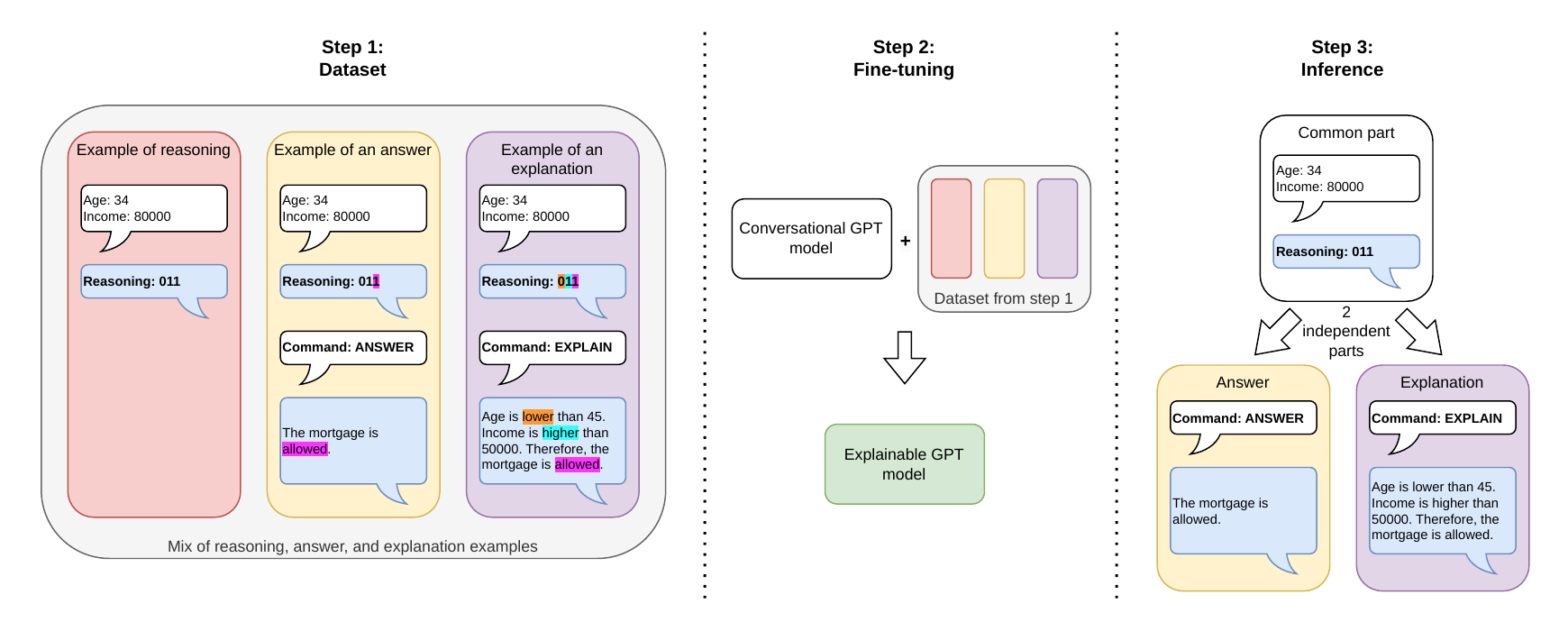}
  }
  \caption{Overview of our methodology. As a first step, we gather a conversational dataset in which for each user input, the triplet of reasoning-answer-explanation ground truths is present. In the second step, we fine-tune a conversational GPT model on the dataset from step 1. As a last step, we perform inference using the fine-tuned model by first computing a reasoning sequence and then including it in the conversation to produce the final answer or explanation, which are obtained independently of each other.}
  \label{fig:overview}
\end{figure*}

We propose to ground natural language explanations, as well as the answers, in a suitable resource-efficient LLM reasoning process. When converted to a sequence of tokens, the result of the reasoning process can then become part of the context observed by the model when producing its final answer or explanation. The reasoning sequence does not have to be directly human-readable, as it merely has to encode the explanation together with the answer. This information can then be simply decoded from the reasoning sequence to natural language when the model generates the final answer or explanation. In order for the explanations to be credible, a joint predict-explain setting can be used, in which the answer and explanation are inferred independently of each other. To demonstrate the plausibility of this approach, we experiment with compact reasoning sequences that we refer to as \textit{compressed chain-of-thought reasoning}. We present the high-level overview of our methodology in Figure \ref{fig:overview}.

We evaluate our explainability framework in an \enquote{LLM-as-a-classifier} setting, in which we train LLMs to mimic the behavior of simple machine learning classifiers such as decision trees. This setting is convenient for our approach as it can be framed so that the final model outputs are affected by multiple intermediate decisions, and also allows for simple and deterministic evaluation of results. Our paper makes the following contributions to the field of LLM explainability:

\begin{enumerate}
    \item We propose an LLM explainability technique for producing faithful natural language explanations grounded in reasoning.
    \item We observe that when a suitable reasoning process is included in LLM training, and the outputs of the reasoning process are placed in the LLM input contexts, LLMs will often copy the partial decisions from the reasoning sequence into their answers or explanations.
    \item We demonstrate the plausibility of our proposed explainability technique by achieving a high alignment between answers and explanations in several problem domains.
    \item We show that besides enabling faithful natural language explanations, the inclusion of the reasoning process can also improve the quality of answers.
\end{enumerate}


\section{Related Work} \label{sec:related-work}

\subsection{LLM Explanations}

LLMs are complex black boxes, and without proper explainability, it is difficult to understand their capabilities, limitations, and potential failures. \cite{zhao2024explainability,weidinger2021ethical}. Explainability techniques are commonly categorized according to several criteria: whether explainability is incorporated into the model's architecture and thus the explanation is part of the model's prediction (\textit{ante-hoc} or \textit{intrinsic explanations}) or if explanations are calculated after the model has been trained and a prediction has been obtained (\textit{post-hoc explanations}); or whether they are related to a single prediction (\textit{local explanations}) or to the general behavior of the model across all predictions (\textit{global explanations}).

\cite{zhao2024explainability} categorizes local LLM explainability techniques into 4 main approaches: feature attribution-based explanations, attention-based explanations, example-based explanations, and natural language explanations.

\subsubsection{Feature Attribution-based Explanations}

These explanations measure the importance of each input feature (such as an input token) in relation to outputs. \textit{Perturbation-based techniques} perturb the inputs using removal or masking \cite{deyoung2019eraser,wu2020perturbed}, which may however generate out-of-distribution data. \textit{Gradient-based} techniques measure partial derivatives of outputs with respect to the input features, using well-established explainability techniques such as \textit{gradient $\times$ input} or \textit{integrated gradients} \cite{sikdar2021integrated,sanyal2021discretized,enguehard2023sequential,hao2021self} which address some of the difficulties that occur when using gradients naively \cite{lyu2024towards}. \textit{Surrogate model methods} employ simpler white-box models to explain individual predictions, notably using the SHAP technique, which utilizes Shapley values and has also been adapted to Transformer models \cite{kokalj2021bert}.

\subsubsection{Attention-based Explanations}

Attention-based explanations analyze the parameters or behavior of attention heads. Numerous studies have focused on explaining attention heads using visualizations, such as with token-level bipartite graphs and heatmaps or neuron-level heatmaps \cite{vig2019multiscale,hoover2019exbert}. Other works have adopted gradient-based methods using various definitions of gradient in attention heads \cite{hao2021self,barkan2021grad}. However, there is ongoing debate on the reliability of attention-based explainability techniques \cite{zhao2024explainability}.

\subsubsection{Example-based Explanations}

These explanations analyze how changes in model inputs affect the outputs. A popular approach is to generate \textit{counterfactual examples} that cause important changes in the outputs by adding, altering, masking, removing, or shuffling words in the input text \cite{wu2021polyjuice}. On the other hand, \textit{adversarial examples} aim to substantially alter the model outputs with barely noticeable changes to the input text \cite{jin2020bert,garg2020bae}. These examples can be added to the training data to improve the robustness of the final model. Another family of methods aims to analyze the impact of the individual training examples on the behavior of the trained model, remarkably without the need for multiple rounds of training \cite{koh2017understanding,yeh2018representer}.

\subsubsection{Natural Language Explanations}

Natural language explanations refer to explanations that take the form of text in natural language, thus making them suitable even for a lay audience \cite{zhao2024explainability,cambria2023survey}. The quality of natural language explanations is commonly assessed according to plausibility, which checks if the explanations are logically sound, faithfulness, which assesses whether the explanations describe the true decision process of the model, and readability \cite{jacovi2020towards}. Although being a relatively large field, most natural language explanation studies focus on other types of models than LLMs. The approaches for LLMs include using simple explain-then-predict and predict-then-answer methods \cite{huang2023can}, training the models using datasets of synthetic \cite{chen2024towards} or human-written explanations \cite{camburu2018snli,rajani2019explain}, and translation of natural language to symbolic solver domains \cite{lyu2023faithful}.

\subsection{LLM Reasoning}

The field of LLM reasoning covers a wide range of methods aimed at improving the model outputs or answers. Chain-of-Thought \cite{kojima2022large} is perhaps the most well-known technique, in which the LLM is simply tasked to reason first before stating the final answer. Extensions of this approach include self-consistency \cite{wang2022self}, in which multiple reasoning paths are sampled, and Tree of Thoughts \cite{yao2023tree}, where the reasoning trajectories form a tree which is explored using search strategies such as BFS and DFS. Other notable reasoning methods include multi-agent collaboration \cite{li2024survey}, knowledge distillation \cite{west2021symbolic}, process-based reward models \cite{lightman2023let}, Monte Carlo Tree Search \cite{feng2023alphazero}, and reinforcement learning \cite{guo2025deepseek}.


\section{Methods} \label{sec:methods}

\subsection{Reasoning-Grounded Natural Language Explanations}

To achieve faithful natural language explainability, we propose to ground LLM explanations as well as answers in a reasoning process. In order to decrease computational complexity, we suggest that the output of the reasoning process does not have to be inherently human-readable, but that it should merely contain the information necessary to be later decoded by the LLM into the final answer or natural language explanation. Such reasoning can be used in a two-step conversational framework, where as the first step, the reasoning sequence is generated, and as the second step, the user (or the chat interface) sends a \textit{command} message indicating whether the model should answer the question or explain the answer, and the model responds accordingly. In case the user chooses to obtain both the answer as well as the explanation, it is crucial that both are obtained independently by the chat interface to prevent the model from fabulating the explanation or the answer being affected by the explanation.

For a clearer definition of the conversational inference process, we can define the conversation history $H_n$ as a sequence of user question messages $U_i$ and model answer messages $A_i$:

\begin{equation}
H_n = U_1 \cdot A_1 \cdot U_2 \cdot A_2 \cdot \ldots \cdot U_{n-1} \cdot A_{n-1} \cdot U_n
\end{equation}

With this notation, the message with reasoning output $R_n$ can be defined by the formula

\begin{equation}
R_n = \text{ReasoningModel}(H_n),
\end{equation}

and the model's answer message $A_n$ and explanation message $E_n$ can be defined by the formulas

\begin{equation}
A_{n} = \text{LLM}(H_n \cdot R_n \cdot C_{\text{answer}})
\end{equation}

and

\begin{equation}
E_{n} = \text{LLM}(H_n \cdot R_n \cdot C_{\text{explain}}),
\end{equation}

where $C_{\text{answer}}$ and $C_{\text{explain}}$ are the \enquote{ANSWER} and \enquote{EXPLAIN} command messages.

As a proof of concept, we experiment with using the LLM to generate compact reasoning sequences in an approach that we refer to as \textit{compressed chain-of-thought reasoning}. With the use of the previous notation, we therefore set $\text{ReasoningModel} = \text{LLM}$. We put three requirements on our compressed chain-of-thought reasoning sequences:

\begin{enumerate}
    \item The reasoning sequences should encode all the partial decisions necessary for the model to produce the right answers.
    \item The reasoning sequences should encode all the partial decisions necessary for the model to produce the natural language explanations in the desired level of detail.
    \item The encoding of the partial decisions in the reasoning sequences should follow a chain-of-thought ordering to allow accurate token-by-token generation by the LLM.
\end{enumerate}

Similarly to regular chain-of-thought reasoning, compressed chain-of-thought reasoning can also potentially improve the quality of LLM answers, as more circuit layer operations can be performed by the LLM before the final answer is produced.

\subsection{LLM as a Classifier}

In our experiments, we adopt an \enquote{LLM-as-a-classifier} approach in which the LLM is tasked with mimicking the behavior of a machine learning classifier. This approach is convenient for our study as reasoning and explanation sequences can be formulated so that they involve chains of various decisions, and we can calculate evaluation metrics deterministically without using methods such as LLM-as-a-judge. We experiment with three problem domains: Logistic regression, decision tree classification, and a natural language dataset generated using decision tree logic. For each problem domain, we design the ground-truth answers to simply state the classification result. For explanations, we use detailed chain-of-thought sequences that describe the intermediate decisions necessary to reach correct classification, and for reasoning, we extract or encode the most important values from the explanation sequence to form minimal, \enquote{compressed} chain-of-thought sequences. For each of the datasets, an example of an input, reasoning, answer, and explanation text for a single instance is shown in Table \ref{table:dataset-instances}.

\begin{table*}[t!]
\caption{Examples of instances from our experimental datasets. For each instance, the sections corresponding to values included in the reasoning sequence are highlighted with bold text, and the occurrences of the ground truth class are underlined.}
\centering
\makebox[\textwidth][c]{
\begin{tabular}{l|p{5cm}p{5cm}p{5cm}}
\toprule
\multirow{2}{*}{} & \multicolumn{3}{c}{\textbf{Dataset}} \\
& \textbf{Logistic regressor} & \textbf{Decision tree} & \textbf{Natural language decision tree} \\
\midrule
Input & \scriptsize{X: [-0.4408, 0.7812, -0.3482, 0.9094, 0.869, -0.0214, -0.0555, -0.8395]} & 
\scriptsize{X: [0.923, 0.252]} & \scriptsize{Loan amount: \$115000.0 Loan-to-value ratio: 92.266 Debt-to-income ratio: <20\% Applicant's age: 25-34 Loan term: 120.0 Income: \$83000.0 Property value: \$475000.0 Total loan costs: \$0.0} \\
\midrule
Reasoning & \scriptsize{\textbf{-1.2465 -1.2465;-2.9536 -4.2001;-2.3885 -6.5886;7.2595 0.6709;-4.5762 -3.9053;0.2138 -3.6915;-0.5065 -4.198;6.8913 2.6933;\underline{1}}} & 
\scriptsize{\textbf{0,0,0,1,1,1,0,\underline{0}}} & \scriptsize{\textbf{1,0,0,0,\underline{1}}} \\
\midrule
Answer & \scriptsize{\textbf{\underline{1}}} & 
\scriptsize{\textbf{\underline{0}}} & \scriptsize{The mortgage is \textbf{\underline{issued}}.} \\
\midrule
Explanation & \scriptsize{[[0, ``w[0] * x[0] = \textbf{-1.2465}'', ``y - 1.2465 = \textbf{-1.2465}''], [1, ``w[1] * x[1] = \textbf{-2.9536}'', ``y - 2.9536 = \textbf{-4.2001}''], [2, ``w[2] * x[2] = \textbf{-2.3885}'', ``y - 2.3885 = \textbf{-6.5886}''], [3, ``w[3] * x[3] = \textbf{7.2595}'', ``y + 7.2595 = \textbf{0.6709}''], [4, ``w[4] * x[4] = \textbf{-4.5762}'', ``y - 4.5762 = \textbf{-3.9053}''], [5, ``w[5] * x[5] = \textbf{0.2138}'', ``y + 0.2138 = \textbf{-3.6915}''], [6, ``w[6] * x[6] = \textbf{-0.5065}'', ``y - 0.5065 = \textbf{-4.198}''], [7, ``w[7] * x[7] = \textbf{6.8913}'', ``y + 6.8913 = \textbf{2.6933}''], [``OUTPUT'', \textbf{\underline{1}}]]} & 
\scriptsize{[[``0.923 < 0.3562'', \textbf{false}], [``0.252 > 0.6825'', \textbf{false}], [``0.923 < 0.5613'', \textbf{false}], [``0.252 < 0.2597'', \textbf{true}], [``0.923 > 0.8087'', \textbf{true}], [``0.252 > 0.0709'', \textbf{true}], [``0.923 < 0.8676'', \textbf{false}], [``OUTPUT'', \textbf{\underline{0}}]]} & \scriptsize{The loan-to-value ratio is \textbf{higher than} 79\%. The income is \textbf{lower than} \$110000. The applicant's age is \textbf{lower or equal to} 34 years. The debt-to-income ratio is \textbf{lower or equal to} 40\%. Therefore, the mortgage is \textbf{\underline{issued}}.} \\
\bottomrule
\end{tabular}
}
\label{table:dataset-instances}
\end{table*}

\subsubsection{Logistic Regression}

In this setting, we randomly generate a parameter vector $\mathbf{w}$ of a logistic regression model without a bias parameter and train the LLM to classify random 8-dimensional input vectors $\mathbf{x}$ according to the following formula:

\begin{equation}y(\mathbf{x}) = \begin{cases}
1& \text{if } \mathbf{w}^T\mathbf{x} > 0\\0 & \text{otherwise.} \end{cases}\end{equation}

\subsubsection{Decision Tree}

In this setting, we randomly generate a binary decision tree of depth 7 with the following node selection logic at each non-leaf node:

\begin{equation}next(N) = \begin{cases}left(N) & \text{if } s_N \times x_{index(N)} > s_N \times t_N\\right(N) & \text{otherwise,}\end{cases}\end{equation}

where $next(N)$ is the next node to be evaluated after the current node $N$, $left(N)$ and $right(N)$ are the left and right child nodes of node $N$, $t_N$ is a random threshold, $index(N)$ is a function that selects the index of $\mathbf{x}$ (defined so that two consecutive nodes can not use the same index), and $s_N$ is a random sign of -1 or 1 that can effectively flip the comparison operator.

Leaf nodes are assigned a class of 0 or 1.

\subsubsection{Decision Tree Encoded in Natural Language}

In the last setting, we experiment with a decision tree that represents a mortgage application review process encoded in natural language. We take a subset of randomly selected mortgage applications from the 2022 version of the \textit{HMDA National Loan Level Dataset} \cite{cfpb2022hmda} as input data and using a manually designed decision tree that represents a fictional mortgage application review process, we generate paragraphs in which each sentence describes a decision branch comparison for one of the input features. Decisions are evaluated for each dataset instance from the top of the decision tree to the leaf with the final class of issued or not issued.


\section{Experiments} \label{sec:experiments}

\subsection{Categorization of Experiments}

\subsubsection{Separate Fine-Tuning for Answers and Explanations}

In this setting, the training dataset is split into two separate datasets, each composed of either input-command-answer or input-command-explanation instances. The LLM is then fine-tuned on each of the two datasets independently, resulting in two fine-tuned models. The inference for answers and explanations is then performed separately using the corresponding model.

\subsubsection{Joint Fine-tuning}

This setting is similar to the previous one, but the answer and explanation instances are not separated into two training datasets. Instead, fine-tuning is performed jointly on the mix of input-command-answer and input-command-explanation examples. Inference is performed in the joint predict-explain approach, with the answers generated independently of the explanations and vice versa, according to the command \enquote{ANSWER} or \enquote{EXPLAIN}.

\subsubsection{Joint Fine-tuning with Reasoning}

In this setting, the training dataset is composed of a mix of input-reasoning, input-reasoning-command-answer, and input-reasoning-command-explanation instances. Inference is performed in two steps, where in the first step, the model generates a reasoning sequence, and in the second step, the answer and explanation are generated independently according to the \enquote{ANSWER} or \enquote{EXPLAIN} command.

\subsubsection{In-context Learning}

To understand how strongly the performance of LLMs is affected by fine-tuning to the specific problem domain, we include results for in-context learning \cite{dong2022survey} as an informative baseline. In this setting, the LLMs are not fine-tuned to the specific classification model, but instead obtain their problem domain knowledge only from classification input-output example pairs included in their input prompt. For each few-shot example, both the answer and explanation target is included. In order to prevent the influence of human prompt engineering, we pre-train the LLMs on a training dataset where each training instance belongs to a different problem domain, corresponding to a randomly generated classifier from the same model family but with different values of model parameters than those used in the test dataset. The input of each few-shot example is randomly generated to achieve greater diversity. We omit the in-context learning setting in the experiments with natural language decision trees due to the complexity of random generation of meaningful decision processes in this problem domain.

\subsection{Experimental Setup}

All experiments were performed using a similar methodology.\footnote{The source code is available under the MIT license at \url{https://github.com/vcahlik/reasoning-grounded-explanations}, together with our datasets.} For each of the five LLMs tested, the model's instruction-tuned variant was used. LLMs were trained using low-rank adaptation \cite{hu2022lora} and Adam optimizer \cite{kingma2014adam} for a single epoch on a train dataset created using 2000 classification inputs. During training, the test loss was periodically measured on a test dataset created using 200 inputs, and at the end of training, the best model checkpoint was kept. In the in-context learning experiments, the number of few-shot examples was 5 for logistic regression and 20 for decision trees. The same training hyperparameters were used in all experiments, namely a batch size of 4 and a learning rate of $5\times10^{-5}$ with a linear schedule and 100 warmup steps.

\subsection{Evaluation Metrics}

In our experiments, we separately measure the classification accuracy of answers and explanations. For explanations, determining the resulting classification is possible as we have designed the explanations as chain-of-thought sequences in which the output class is always stated at the end. Furthermore, we measure the rate of alignment between the answer and explanation classifications.

\section{Results}

\subsection{Logistic Regression Results}

The results for the logistic regression dataset are shown in Table \ref{table:logistic-regression}. In-context learning has near-perfect classification accuracy for explanations, as the parameters of the logistic regressor are stated in the few-shot examples and therefore it is simple for the model to generate correct chain-of-thought explanations. However, for answers, classification accuracy is equivalent to random guessing due to the difficulty of the task when a chain-of-thought process is not used. The gap between the classification accuracy for answers and explanations is also wide for most of the fine-tuning experiments without reasoning. However, when reasoning is used, the classification accuracies of answers increase to the level of classification accuracies for explanations, indicating that the reasoning process helps the LLMs achieve correct answers.

\begin{table*}
\caption{Results on the logistic regression dataset. The experimental setup differs in whether training was performed separately or jointly for answers and explanations, whether in-context learning (ICL) was used, and whether reasoning was used. Outputs that could not be parsed into a valid class are counted towards errors and their rate is additionally shown in italic.}
\centering
\makebox[\textwidth][c]{
\begin{tabular}{llll|ccccc}
\toprule
\textbf{Ans./exp. training} & \textbf{ICL} & \textbf{Reasoning} & \textbf{Metric} & \textbf{Llama 3 8B} & \textbf{Mistral NeMo} & \textbf{Mistral 7B} & \textbf{Zephyr SFT} & \textbf{Phi-4} \\
\midrule
\multirow{3}{*}{Separately} & \multirow{3}{*}{Yes} & \multirow{3}{*}{No} & Answer acc. & 0.455 & 0.515 & 0.470 & 0.495 & 0.470 \\
& & & Explanation acc. & 0.990 & 1.000 & 0.995 & 1.000 & 0.990 \\
& & & Alignment rate & 0.455 & 0.515 & 0.475 & 0.495 & 0.460 \\\midrule
\multirow{3}{*}{Separately} & \multirow{3}{*}{No} & \multirow{3}{*}{No} & Answer acc. & 0.610 & 0.890 & 0.830 & 0.555 & 0.620 \\
& & & Explanation acc. & 0.615 \textit{(0.140)} & 0.990 \textit{(0.005)} & 0.995 & 0.995 & 0.990 \\
& & & Alignment rate & 0.455 \textit{(0.140)} & 0.880 \textit{(0.005)} & 0.835 & 0.560 & 0.620 \\\midrule
\multirow{3}{*}{Jointly} & \multirow{3}{*}{No} & \multirow{3}{*}{No} & Answer acc. & 0.640 & 0.530 & 0.555 & 0.470 & 0.470 \\
& & & Explanation acc. & 0.990 & 1.000 & 1.000 & 0.995 & 1.000 \\
& & & Alignment rate & 0.630 & 0.530 & 0.555 & 0.475 & 0.470 \\\midrule
\multirow{3}{*}{Jointly} & \multirow{3}{*}{No} & \multirow{3}{*}{Yes} & Answer acc. & 0.890 \textit{(0.020)} & 0.995 & 1.000 & 1.000 & 0.945 \\
& & & Explanation acc. & 0.875 \textit{(0.030)} & 0.995 & 1.000 & 1.000 & 0.940 \textit{(0.005)} \\
& & & Alignment rate & 0.965 \textit{(0.035)} & 1.000 & 1.000 & 1.000 & 0.995 \textit{(0.005)} \\\midrule
\bottomrule
\end{tabular}
}
\label{table:logistic-regression}
\end{table*}

\subsection{Decision Tree Results}

\begin{figure}[t!]
    \centering
    \makebox[\textwidth][c]{ 
        \begin{subfigure}{0.7\textwidth}
            \centering
            \includegraphics[width=\linewidth]{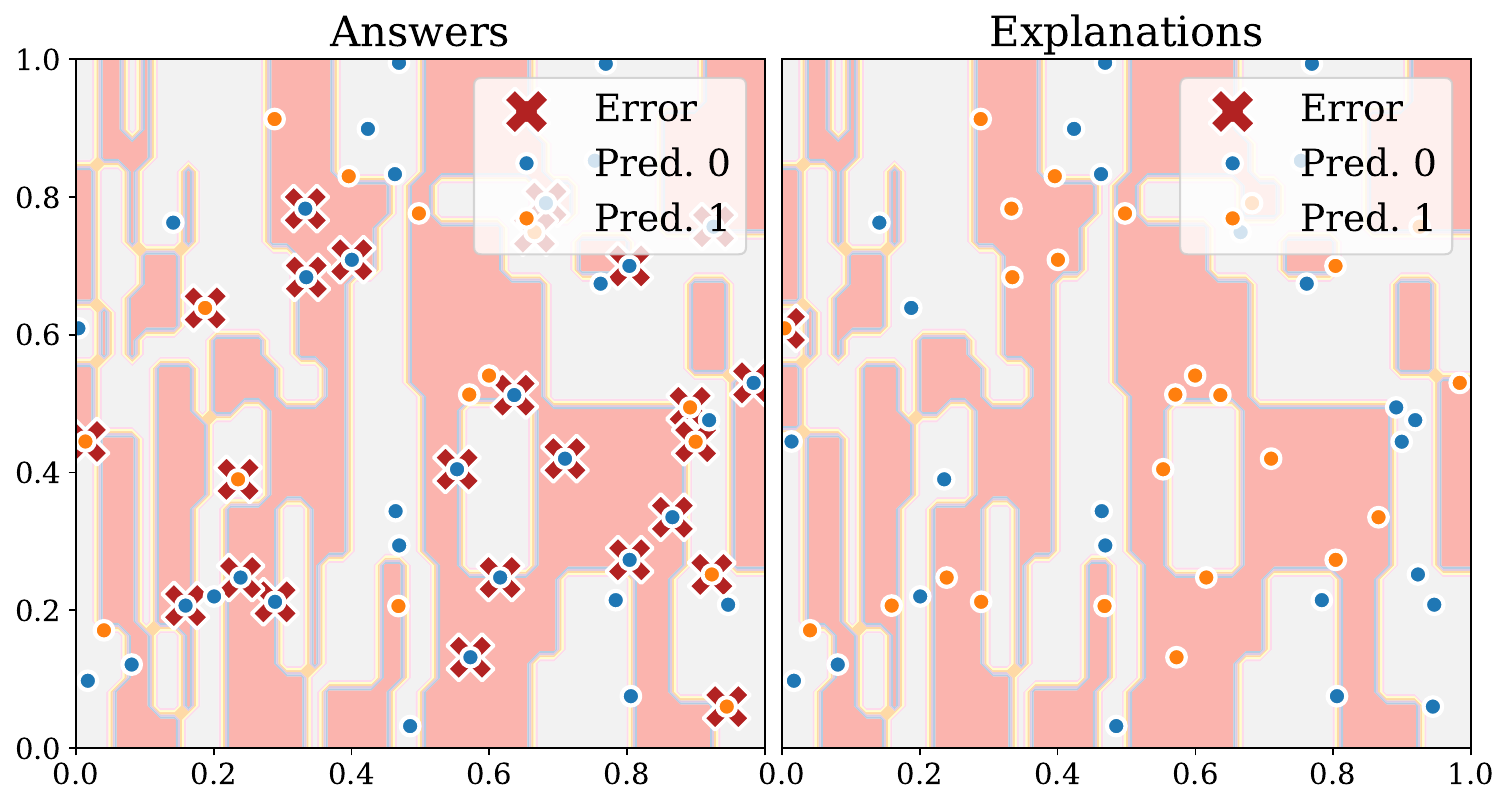}
            \caption{Results without reasoning}
            \label{fig:without-reasoning}
        \end{subfigure}%
        \hspace{0.025\textwidth} 
        \begin{subfigure}{0.7\textwidth}
            \centering
            \includegraphics[width=\linewidth]{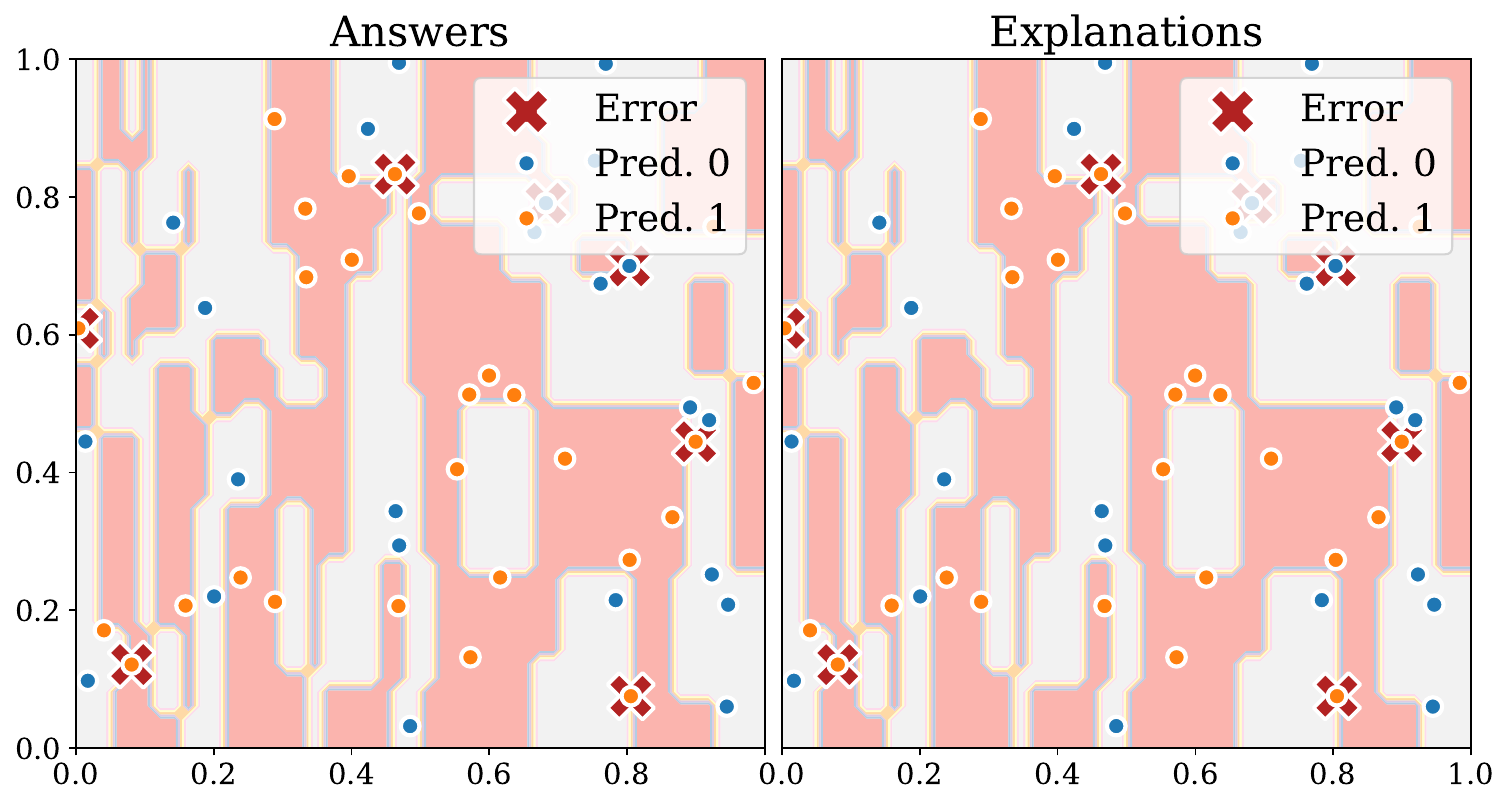}
            \caption{Results with reasoning}
            \label{fig:with-reasoning}
        \end{subfigure}
    }
    \caption{Experiments with joint training of answers and explanations on a decision tree dataset. The colored regions correspond to ground-truth classes. When reasoning is used, answer and explanation classification errors are typically near-perfectly aligned.}
    \label{fig:decision-tree-plots}
\end{figure}

The results for the decision tree dataset, belonging to a tree of depth 7, are shown in Table \ref{table:decision-trees}. Even though 4 times as many few-shot examples were used than in the logistic regression experiments, the classification accuracy of in-context learning is low even for explanations, as the number of few-shot examples is still lower than the number of decision tree leaves. The results for fine-tuning without reasoning are similar to those with the logistic regression dataset. Results for reasoning show higher error rates than in the logistic regression experiments, presumably due to the less detailed reasoning process. However, answers and explanations now remarkably contain the same classification errors in almost all cases, as can be seen from the near-perfect alignment rates. We visualize this phenomenon by plotting the classifications for one of the experiments in Figure \ref{fig:decision-tree-plots}.

\begin{figure*}[b!]
  \centering
  \makebox[\textwidth][c]{
    \includegraphics[width=1\textwidth]{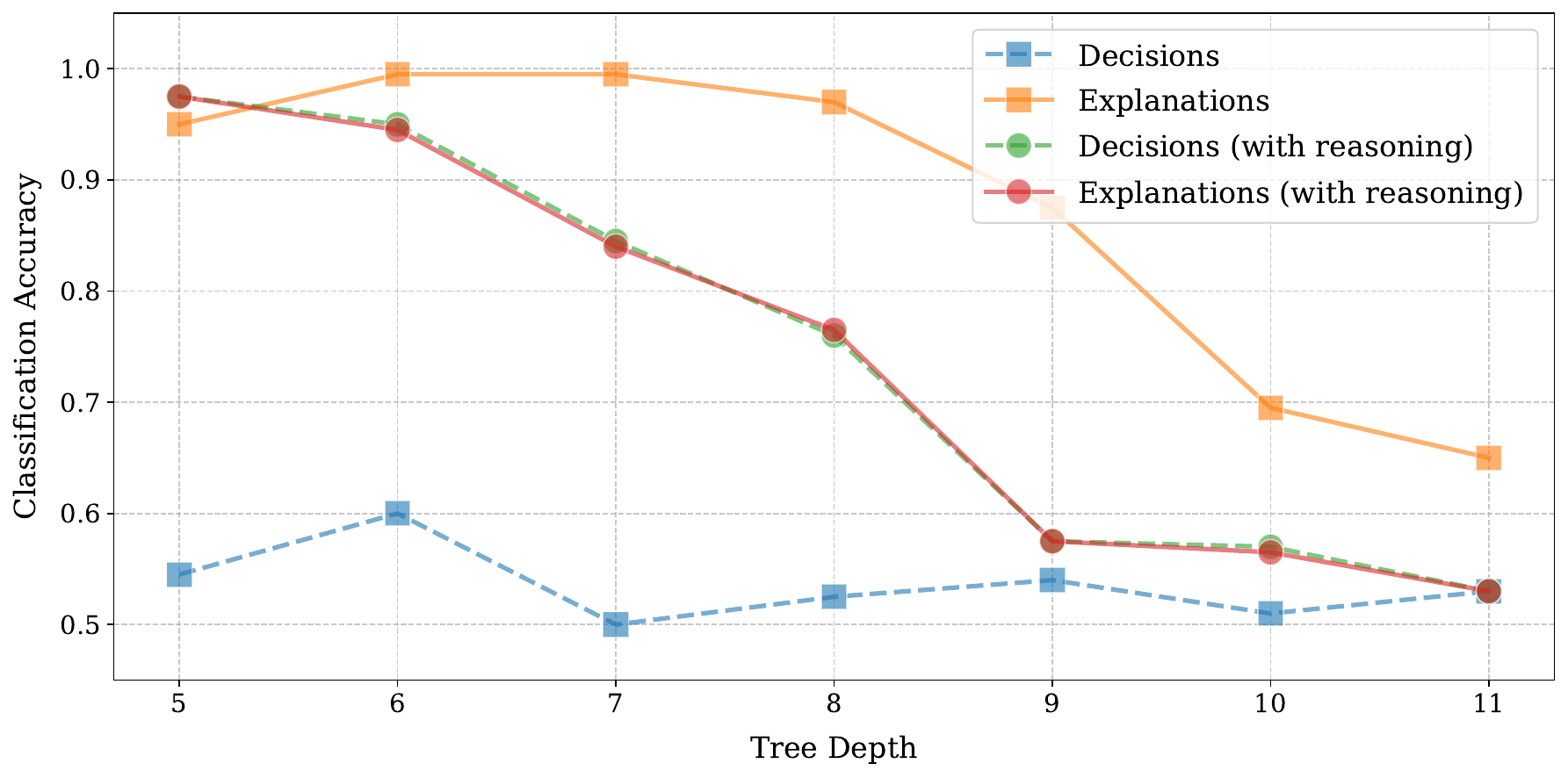}
  }
  \vskip-0.cm
  \caption{Classification accuracies for experiments with decision trees of various depths}
  \label{fig:tree-depths}
\end{figure*}

Figure \ref{fig:tree-depths} shows the results for the Mistral 7B model on datasets of varying decision tree depths, indicating that classification accuracy tends to decrease as the complexity of the trees increases. With reasoning, answer and explanation classifications are near-perfectly aligned for all of the depths, in contrast to the alignment rates for experiments without reasoning. However, explanations without reasoning tend to have the highest classification accuracy in these experiments, supposedly due to the chain-of-thought explanation sequences being more thorough than the compressed chain-of-thought reasoning sequences.

\begin{table*}
\caption{Results on the decision tree dataset, with the same semantics as in Table \ref{table:logistic-regression}}
\centering
\makebox[\textwidth][c]{
\begin{tabular}{llll|ccccc}
\toprule
\textbf{Ans./exp. training} & \textbf{ICL} & \textbf{Reasoning} & \textbf{Metric} & \textbf{Llama 3 8B} & \textbf{Mistral NeMo} & \textbf{Mistral 7B} & \textbf{Zephyr SFT} & \textbf{Phi-4} \\
\midrule
\multirow{3}{*}{Separately} & \multirow{3}{*}{Yes} & \multirow{3}{*}{No} & Answer acc. & 0.535 & 0.510 & 0.490 & 0.490 & 0.535 \\
& & & Explanation acc. & 0.670 & 0.685 & 0.695 & 0.695 & 0.700 \\
& & & Alignment rate & 0.565 & 0.565 & 0.495 & 0.485 & 0.505 \\\midrule
\multirow{3}{*}{Separately} & \multirow{3}{*}{No} & \multirow{3}{*}{No} & Answer acc. & 0.475 & 0.525 & 0.530 & 0.565 & 0.565 \\
& & & Explanation acc. & 0.975 & 0.985 & 0.985 & 1.000 & 0.955 \\
& & & Alignment rate & 0.480 & 0.540 & 0.515 & 0.565 & 0.580 \\\midrule
\multirow{3}{*}{Jointly} & \multirow{3}{*}{No} & \multirow{3}{*}{No} & Answer acc. & 0.475 & 0.450 & 0.500 & 0.475 & 0.520 \\
& & & Explanation acc. & 0.985 & 0.985 & 0.995 & 0.975 & 0.985 \\
& & & Alignment rate & 0.480 & 0.435 & 0.505 & 0.460 & 0.505 \\\midrule
\multirow{3}{*}{Jointly} & \multirow{3}{*}{No} & \multirow{3}{*}{Yes} & Answer acc. & 0.745 & 0.835 & 0.845 & 0.875 & 0.715 \\
& & & Explanation acc. & 0.745 & 0.835 & 0.840 \textit{(0.005)} & 0.875 & 0.715 \\
& & & Alignment rate & 1.000 & 1.000 & 0.995 \textit{(0.005)} & 1.000 & 1.000 \\\midrule
\bottomrule
\end{tabular}
}
\label{table:decision-trees}
\end{table*}

\subsection{Natural Language Decision Tree Results}

The results for the natural language decision tree dataset, shown in Table \ref{table:natural-language-decision-trees}, are similar to those for the decision tree dataset. However, in this case, the use of reasoning is associated with near-perfect classification accuracies for both answers and explanations and with perfect alignment of classification errors.

\begin{table*}
\caption{Results on the natural language decision tree dataset, with the same semantics as in Table \ref{table:logistic-regression}}
\centering
\makebox[\textwidth][c]{
\begin{tabular}{llll|ccccc}
\toprule
\textbf{Ans./exp. training} & \textbf{ICL} & \textbf{Reasoning} & \textbf{Metric} & \textbf{Llama 3 8B} & \textbf{Mistral NeMo} & \textbf{Mistral 7B} & \textbf{Zephyr SFT} & \textbf{Phi-4} \\
\midrule
\multirow{3}{*}{Separately} & \multirow{3}{*}{No} & \multirow{3}{*}{No} & Answer acc. & 0.810 & 0.825 & 0.850 & 0.850 & 0.815 \\
& & & Explanation acc. & 0.950 & 0.975 & 0.950 & 0.995 & 0.970 \\
& & & Alignment rate & 0.860 & 0.840 & 0.850 & 0.855 & 0.835 \\\midrule
\multirow{3}{*}{Jointly} & \multirow{3}{*}{No} & \multirow{3}{*}{No} & Answer acc. & 0.840 & 0.840 & 0.810 & 0.865 & 0.845 \\
& & & Explanation acc. & 0.935 & 0.975 & 0.980 & 0.990 & 0.975 \\
& & & Alignment rate & 0.825 & 0.855 & 0.820 & 0.875 & 0.830 \\\midrule
\multirow{3}{*}{Jointly} & \multirow{3}{*}{No} & \multirow{3}{*}{Yes} & Answer acc. & 0.985 & 0.970 & 0.985 & 1.000 & 1.000 \\
& & & Explanation acc. & 0.985 & 0.970 & 0.985 & 1.000 & 1.000 \\
& & & Alignment rate & 1.000 & 1.000 & 1.000 & 1.000 & 1.000 \\\midrule
\bottomrule
\end{tabular}
}
\label{table:natural-language-decision-trees}
\end{table*}

\subsection{Analysis of Errors}

As a further analysis, we study the partial decisions present in the reasoning and explanation sequences generated on the decision tree dataset by the Llama 3 8B model. As shown in Table \ref{table:error-analysis}, the final classification errors in both the reasoning and explanation sequences are caused by the accumulation of mistakes in partial decisions. It is noteworthy that all of the decisions are perfectly aligned between the reasoning and explanation sequences. Although not shown in the table, we also observed perfect alignment between the answer classifications and reasoning classifications, meaning that all partial decisions as well as the final classifications are aligned between the generated answers, explanations, and reasoning in this case.

\begin{table*}[b!]
\caption{Analysis of the correctness of the reasoning and explanation chain-of-thought sequences generated by Llama 3 8B on the decision tree dataset}
\centering
\makebox[\textwidth][c]{
\begin{tabular}{l|cccccccc}
\toprule
\multirow{2}{*}{} & \multicolumn{7}{c}{\textbf{Partial decision}} & \multirow{2}{*}{\textbf{Final classification}} \\
& \textbf{1} & \textbf{2} & \textbf{3} & \textbf{4} & \textbf{5} & \textbf{6} & \textbf{7} & \\
\midrule
Reasoning accuracy & 0.995 & 1.000 & 0.990 & 0.960 & 0.895 & 0.830 & 0.745 & 0.745 \\
Explanation accuracy & 0.995 & 1.000 & 0.990 & 0.960 & 0.895 & 0.830 & 0.745 & 0.745 \\
Alignment rate & 1.000 & 1.000 & 1.000 & 1.000 & 1.000 & 1.000 & 1.000 & 1.000 \\
\bottomrule
\end{tabular}
}
\label{table:error-analysis}
\end{table*}


\section{Discussion} \label{sec:discussion}

It may not be immediately clear why the inclusion of reasoning sequences in LLM input contexts leads to alignment between answers and explanations. It seems that during training, the LLM must learn the relatively simple task of reproducing the compressed chain-of-thought reasoning sequence to succeed at the more difficult task of producing the one-step answer classifications. We hypothesize that once the model learns to produce accurate reasoning sequences, the internal mechanism by which the LLM produces its explanations also degrades to the copying of the partial decisions from the reasoning sequence. To gain supporting evidence for this hypothesis, we further experimented with randomly flipping the partial decisions as well as the final classification decisions present in the reasoning sequences produced by fine-tuned Llama 3 8B. As was suspected, we observed that almost all of the changes were propagated into the produced explanations as well as to the answers.

Our approach presented in this paper can be extended in numerous possible ways, which we leave for future work. Primarily, the reasoning process that we chose for our proof-of-concept experiments could be extended to wider problem domains or even to general-purpose assistant datasets, for example by straightforwardly using chain-of-thought reasoning or similar approaches, such as the reasoning process used by DeepSeek-r1 \cite{guo2025deepseek}. It would also seem beneficial to introduce a training loss that directly penalizes the mismatch between answers, explanations, and reasoning. Furthermore, we believe that LLM applications could benefit from other output modes besides answering and explaining. We envision a multitask setting with additional implemented commands, such as those for obtaining explanations of varying detail, classification of user intent, content filtering analysis, metadata generation, and so on.


\section{Conclusion} \label{sec:conclusion}

In this paper, we have proposed an LLM explainability technique for obtaining faithful natural language explanations by grounding the LLM answers and explanations in a reasoning process. We have shown that LLMs often simply copy the partial decisions from the reasoning sequence into their answers or explanations, and we utilized this phenomenon to achieve high alignment between answers and explanations in several problem domains. Furthermore, we have shown that besides enabling faithful explanations, the use of a reasoning process can also lead to improvements in the quality of answers. We hope that our study inspires further research or real-world use-cases that advance the current state of explainability in LLMs.


\begin{credits}
\subsubsection{\ackname} The access to the computational infrastructure of the OP VVV funded project CZ.02.1.01/0.0/0.0/16\_019/0000765 ``Research Center for Informatics'' is gratefully acknowledged.

\subsubsection{\discintname}
The authors have no competing interests to declare that are relevant to the content of this article.
\end{credits}


\bibliographystyle{unsrt}
\bibliography{bibliography}

@article{zhao2024explainability,
  title={Explainability for large language models: A survey},
  author={Zhao, Haiyan and Chen, Hanjie and Yang, Fan and Liu, Ninghao and Deng, Huiqi and Cai, Hengyi and Wang, Shuaiqiang and Yin, Dawei and Du, Mengnan},
  journal={ACM Transactions on Intelligent Systems and Technology},
  volume={15},
  number={2},
  pages={1--38},
  year={2024},
  publisher={ACM New York, NY}
}

@article{weidinger2021ethical,
  title={Ethical and social risks of harm from language models},
  author={Weidinger, Laura and Mellor, John and Rauh, Maribeth and Griffin, Conor and Uesato, Jonathan and Huang, Po-Sen and Cheng, Myra and Glaese, Mia and Balle, Borja and Kasirzadeh, Atoosa and others},
  journal={arXiv preprint arXiv:2112.04359},
  year={2021}
}

@article{wu2020perturbed,
  title={Perturbed masking: Parameter-free probing for analyzing and interpreting BERT},
  author={Wu, Zhiyong and Chen, Yun and Kao, Ben and Liu, Qun},
  journal={arXiv preprint arXiv:2004.14786},
  year={2020}
}

@article{deyoung2019eraser,
  title={ERASER: A benchmark to evaluate rationalized NLP models},
  author={DeYoung, Jay and Jain, Sarthak and Rajani, Nazneen Fatema and Lehman, Eric and Xiong, Caiming and Socher, Richard and Wallace, Byron C},
  journal={arXiv preprint arXiv:1911.03429},
  year={2019}
}

@article{lyu2024towards,
  title={Towards faithful model explanation in nlp: A survey},
  author={Lyu, Qing and Apidianaki, Marianna and Callison-Burch, Chris},
  journal={Computational Linguistics},
  pages={1--67},
  year={2024},
  publisher={MIT Press 255 Main Street, 9th Floor, Cambridge, Massachusetts 02142, USA~…}
}

@article{enguehard2023sequential,
  title={Sequential Integrated Gradients: a simple but effective method for explaining language models},
  author={Enguehard, Joseph},
  journal={arXiv preprint arXiv:2305.15853},
  year={2023}
}

@inproceedings{sikdar2021integrated,
  title={Integrated directional gradients: Feature interaction attribution for neural NLP models},
  author={Sikdar, Sandipan and Bhattacharya, Parantapa and Heese, Kieran},
  booktitle={Proceedings of the 59th Annual Meeting of the Association for Computational Linguistics and the 11th International Joint Conference on Natural Language Processing (Volume 1: Long Papers)},
  pages={865--878},
  year={2021}
}

@article{sanyal2021discretized,
  title={Discretized integrated gradients for explaining language models},
  author={Sanyal, Soumya and Ren, Xiang},
  journal={arXiv preprint arXiv:2108.13654},
  year={2021}
}

@inproceedings{hao2021self,
  title={Self-attention attribution: Interpreting information interactions inside transformer},
  author={Hao, Yaru and Dong, Li and Wei, Furu and Xu, Ke},
  booktitle={Proceedings of the AAAI Conference on Artificial Intelligence},
  volume={35},
  number={14},
  pages={12963--12971},
  year={2021}
}

@inproceedings{kokalj2021bert,
  title={BERT meets shapley: Extending SHAP explanations to transformer-based classifiers},
  author={Kokalj, Enja and {\v{S}}krlj, Bla{\v{z}} and Lavra{\v{c}}, Nada and Pollak, Senja and Robnik-{\v{S}}ikonja, Marko},
  booktitle={Proceedings of the EACL hackashop on news media content analysis and automated report generation},
  pages={16--21},
  year={2021}
}

@article{vig2019multiscale,
  title={A multiscale visualization of attention in the transformer model},
  author={Vig, Jesse},
  journal={arXiv preprint arXiv:1906.05714},
  year={2019}
}

@article{hoover2019exbert,
  title={exbert: A visual analysis tool to explore learned representations in transformers models},
  author={Hoover, Benjamin and Strobelt, Hendrik and Gehrmann, Sebastian},
  journal={arXiv preprint arXiv:1910.05276},
  year={2019}
}

@inproceedings{barkan2021grad,
  title={Grad-sam: Explaining transformers via gradient self-attention maps},
  author={Barkan, Oren and Hauon, Edan and Caciularu, Avi and Katz, Ori and Malkiel, Itzik and Armstrong, Omri and Koenigstein, Noam},
  booktitle={Proceedings of the 30th ACM International Conference on Information \& Knowledge Management},
  pages={2882--2887},
  year={2021}
}

@inproceedings{jin2020bert,
  title={Is bert really robust? a strong baseline for natural language attack on text classification and entailment},
  author={Jin, Di and Jin, Zhijing and Zhou, Joey Tianyi and Szolovits, Peter},
  booktitle={Proceedings of the AAAI conference on artificial intelligence},
  volume={34},
  number={05},
  pages={8018--8025},
  year={2020}
}

@article{garg2020bae,
  title={Bae: Bert-based adversarial examples for text classification},
  author={Garg, Siddhant and Ramakrishnan, Goutham},
  journal={arXiv preprint arXiv:2004.01970},
  year={2020}
}

@article{wu2021polyjuice,
  title={Polyjuice: Generating counterfactuals for explaining, evaluating, and improving models},
  author={Wu, Tongshuang and Ribeiro, Marco Tulio and Heer, Jeffrey and Weld, Daniel S},
  journal={arXiv preprint arXiv:2101.00288},
  year={2021}
}

@article{yeh2018representer,
  title={Representer point selection for explaining deep neural networks},
  author={Yeh, Chih-Kuan and Kim, Joon and Yen, Ian En-Hsu and Ravikumar, Pradeep K},
  journal={Advances in neural information processing systems},
  volume={31},
  year={2018}
}

@inproceedings{koh2017understanding,
  title={Understanding black-box predictions via influence functions},
  author={Koh, Pang Wei and Liang, Percy},
  booktitle={International conference on machine learning},
  pages={1885--1894},
  year={2017},
  organization={PMLR}
}

@book{russell2019human,
  title={Human compatible: AI and the problem of control},
  author={Russell, Stuart},
  year={2019},
  publisher={Penguin Uk}
}

@article{dong2022survey,
  title={A survey on in-context learning},
  author={Dong, Qingxiu and Li, Lei and Dai, Damai and Zheng, Ce and Ma, Jingyuan and Li, Rui and Xia, Heming and Xu, Jingjing and Wu, Zhiyong and Liu, Tianyu and others},
  journal={arXiv preprint arXiv:2301.00234},
  year={2022}
}

@article{hu2022lora,
  title={Lora: Low-rank adaptation of large language models.},
  author={Hu, Edward J and Shen, Yelong and Wallis, Phillip and Allen-Zhu, Zeyuan and Li, Yuanzhi and Wang, Shean and Wang, Lu and Chen, Weizhu and others},
  journal={ICLR},
  volume={1},
  number={2},
  pages={3},
  year={2022}
}

@article{kingma2014adam,
  title={Adam: A method for stochastic optimization},
  author={Kingma, Diederik P and Ba, Jimmy},
  journal={arXiv preprint arXiv:1412.6980},
  year={2014}
}

@article{chen2024towards,
  title={Towards consistent natural-language explanations via explanation-consistency finetuning},
  author={Chen, Yanda and Singh, Chandan and Liu, Xiaodong and Zuo, Simiao and Yu, Bin and He, He and Gao, Jianfeng},
  journal={arXiv preprint arXiv:2401.13986},
  year={2024}
}

@article{cambria2023survey,
  title={A survey on XAI and natural language explanations},
  author={Cambria, Erik and Malandri, Lorenzo and Mercorio, Fabio and Mezzanzanica, Mario and Nobani, Navid},
  journal={Information Processing \& Management},
  volume={60},
  number={1},
  pages={103111},
  year={2023},
  publisher={Elsevier}
}

@article{guo2025deepseek,
  title={Deepseek-r1: Incentivizing reasoning capability in llms via reinforcement learning},
  author={Guo, Daya and Yang, Dejian and Zhang, Haowei and Song, Junxiao and Zhang, Ruoyu and Xu, Runxin and Zhu, Qihao and Ma, Shirong and Wang, Peiyi and Bi, Xiao and others},
  journal={arXiv preprint arXiv:2501.12948},
  year={2025}
}

@misc{cfpb2022hmda,
  author       = {{Consumer Financial Protection Bureau}},
  title        = {{HMDA National Loan Level Dataset 2022}},
  year         = {2022},
  publisher    = {Consumer Financial Protection Bureau},
  url          = {https://ffiec.cfpb.gov/data-publication/snapshot-national-loan-level-dataset/2022}
}

@article{kojima2022large,
  title={Large language models are zero-shot reasoners},
  author={Kojima, Takeshi and Gu, Shixiang Shane and Reid, Machel and Matsuo, Yutaka and Iwasawa, Yusuke},
  journal={Advances in neural information processing systems},
  volume={35},
  pages={22199--22213},
  year={2022}
}

@article{wang2022self,
  title={Self-consistency improves chain of thought reasoning in language models},
  author={Wang, Xuezhi and Wei, Jason and Schuurmans, Dale and Le, Quoc and Chi, Ed and Narang, Sharan and Chowdhery, Aakanksha and Zhou, Denny},
  journal={arXiv preprint arXiv:2203.11171},
  year={2022}
}

@article{yao2023tree,
  title={Tree of thoughts: Deliberate problem solving with large language models},
  author={Yao, Shunyu and Yu, Dian and Zhao, Jeffrey and Shafran, Izhak and Griffiths, Tom and Cao, Yuan and Narasimhan, Karthik},
  journal={Advances in neural information processing systems},
  volume={36},
  pages={11809--11822},
  year={2023}
}

@article{li2024survey,
  title={A survey on LLM-based multi-agent systems: workflow, infrastructure, and challenges},
  author={Li, Xinyi and Wang, Sai and Zeng, Siqi and Wu, Yu and Yang, Yi},
  journal={Vicinagearth},
  volume={1},
  number={1},
  pages={9},
  year={2024},
  publisher={Springer}
}

@article{west2021symbolic,
  title={Symbolic knowledge distillation: from general language models to commonsense models},
  author={West, Peter and Bhagavatula, Chandra and Hessel, Jack and Hwang, Jena D and Jiang, Liwei and Bras, Ronan Le and Lu, Ximing and Welleck, Sean and Choi, Yejin},
  journal={arXiv preprint arXiv:2110.07178},
  year={2021}
}

@inproceedings{lightman2023let,
  title={Let's verify step by step},
  author={Lightman, Hunter and Kosaraju, Vineet and Burda, Yuri and Edwards, Harrison and Baker, Bowen and Lee, Teddy and Leike, Jan and Schulman, John and Sutskever, Ilya and Cobbe, Karl},
  booktitle={The Twelfth International Conference on Learning Representations},
  year={2023}
}

@article{feng2023alphazero,
  title={Alphazero-like tree-search can guide large language model decoding and training},
  author={Feng, Xidong and Wan, Ziyu and Wen, Muning and McAleer, Stephen Marcus and Wen, Ying and Zhang, Weinan and Wang, Jun},
  journal={arXiv preprint arXiv:2309.17179},
  year={2023}
}

@article{jacovi2020towards,
  title={Towards faithfully interpretable NLP systems: How should we define and evaluate faithfulness?},
  author={Jacovi, Alon and Goldberg, Yoav},
  journal={arXiv preprint arXiv:2004.03685},
  year={2020}
}

@article{camburu2018snli,
  title={e-snli: Natural language inference with natural language explanations},
  author={Camburu, Oana-Maria and Rockt{\"a}schel, Tim and Lukasiewicz, Thomas and Blunsom, Phil},
  journal={Advances in Neural Information Processing Systems},
  volume={31},
  year={2018}
}

@article{rajani2019explain,
  title={Explain yourself! leveraging language models for commonsense reasoning},
  author={Rajani, Nazneen Fatema and McCann, Bryan and Xiong, Caiming and Socher, Richard},
  journal={arXiv preprint arXiv:1906.02361},
  year={2019}
}

@inproceedings{lyu2023faithful,
  title={Faithful chain-of-thought reasoning},
  author={Lyu, Qing and Havaldar, Shreya and Stein, Adam and Zhang, Li and Rao, Delip and Wong, Eric and Apidianaki, Marianna and Callison-Burch, Chris},
  booktitle={The 13th International Joint Conference on Natural Language Processing and the 3rd Conference of the Asia-Pacific Chapter of the Association for Computational Linguistics (IJCNLP-AACL 2023)},
  year={2023}
}

@article{huang2023can,
  title={Can large language models explain themselves? a study of llm-generated self-explanations},
  author={Huang, Shiyuan and Mamidanna, Siddarth and Jangam, Shreedhar and Zhou, Yilun and Gilpin, Leilani H},
  journal={arXiv preprint arXiv:2310.11207},
  year={2023}
}

\end{document}